%% file: spill_crumbs_wiping.ICRA23.tex
\documentclass[letterpaper, 10 pt, conference]{ieeeconf}  %

\IEEEoverridecommandlockouts                              %

\input{preamble.tex}

\title{Robotic Table Wiping via Reinforcement Learning and \\Whole-body Trajectory Optimization}

\author{Thomas Lew$^{1,3}$, Sumeet Singh$^{1}$, Mario Prats$^{2}$, Jeffrey Bingham$^{2}$, Jonathan Weisz$^{2}$, Benjie Holson$^{2}$, Xiaohan \\
Zhang$^{1,4}$, Vikas Sindhwani$^{1}$, Yao Lu$^{1}$, Fei Xia$^{1}$, Peng Xu$^{1}$, Tingnan Zhang$^{1}$, Jie Tan$^{1}$, Montserrat Gonzalez$^{1}$%
\thanks{$^{1}$Robotics at Google \quad $^{2}$\href{www.everydayrobots.com}{Everyday Robots}}%
\thanks{$^{3}$Department of Aeronautics and Astronautics, Stanford University}%
\thanks{$^{4}$Department of Computer Science, SUNY Binghamton}%
}

\begin{document}

\maketitle
 \thispagestyle{empty}
\pagestyle{empty}

\begin{abstract}
We propose a framework to enable multipurpose assistive mobile robots to autonomously wipe tables to clean spills and crumbs. This problem is challenging, as it requires planning wiping actions while reasoning over uncertain latent dynamics of crumbs and spills captured via high-dimensional visual observations. Simultaneously, we must guarantee constraints satisfaction to enable safe deployment in unstructured cluttered environments.   
To tackle this problem, we first propose a stochastic differential equation to model crumbs and spill dynamics and absorption with a robot wiper. 
Using this model, we train a vision-based policy for planning wiping actions in simulation using reinforcement learning (RL). 
To enable zero-shot sim-to-real deployment, we dovetail the RL policy with a whole-body trajectory optimization framework to compute base and arm joint trajectories that execute the desired wiping motions while guaranteeing constraints satisfaction. 
We extensively validate our approach in simulation and on hardware.
\\[1mm]
Video of experiments: \scalebox{0.95}{\url{https://youtu.be/inORKP4F3EI}}
\end{abstract}

\section{Introduction}\label{sec:introduction}
Multipurpose assistive robots will play an important role in improving people's lives in the spaces where we live and work \cite{Kim2019,Bajracharya2020}. Repetitive tasks such as cleaning surfaces are well-suited for robots, but remain challenging for systems that typically operate in structured environments. 
Operating in the real world requires handling high-dimensional sensory inputs and dealing with the stochasticity of the environment. 

Learning-based techniques such as reinforcement learning (RL) offer the promise of solving these complex visuo-motor tasks from high-dimensional observations. 
However, applying end-to-end learning methods to mobile manipulation tasks remains challenging due to the increased dimensionality and the need for precise low-level control. 
Additionally, on-robot deployment either requires collecting large amounts of data \cite{Kalashnikov2018,James2019,Rao2020}, using accurate but computationally expensive models \cite{Rao2020}, %
or on-hardware fine-tuning \cite{Ghadirzadeh2021}. %

In this work, we focus on the task of cleaning tables with a mobile robotic manipulator equipped with a wiping tool. This problem is challenging for both high-level planning and low-level control. 
    Indeed, at a high-level, deciding how to best wipe a spill perceived by a camera requires solving a challenging planning problem with stochastic dynamics. 
    At a low-level, executing a wiping motion requires simultaneously maintaining contact with the table while avoiding nearby obstacles such as chairs. Designing a real-time and effective solution to this problem remains an open problem \cite{Kim2019}. 
    
    Our main contributions are as follows:
\begin{itemize}[leftmargin=4.5mm]
    \item We propose a framework for autonomous table wiping. First, we use visual observations of the table state to plan high-level wiping actions for the end-effector. Then, we compute whole-body trajectories that we execute using admittance control. This approach is key to achieving reliable table wiping in new environments without the need for real-world data collection or demonstrations.
    \item We describe the uncertain time evolution of dirty particles on the table using a stochastic differential equation (SDE) capable of modeling absorption with the wiper. 
    Then, we formulate the problem of planning wiping actions as a stochastic optimal control problem. %
    As this task requires planning over high-dimensional visual inputs, we solve the problem using RL, \emph{entirely in simulation}. 
    \item We design a whole-body trajectory optimization algorithm %
    for navigation in cluttered environments and table wiping. Our approach accounts for the kinematics of the manipulator, the nonholonomic constraints of the base, and collision avoidance constraints with the environment.  %
\end{itemize}
This approach combines the strengths of reinforcement learning - planning in high-dimensional observation spaces with complex stochastic dynamics, and of trajectory optimization - guaranteeing constraints satisfaction while executing whole-body trajectories; it does not require collecting a task-specific dataset on the system, and transfers zero-shot to hardware. %

\begin{figure}[!t]%
        \centering
        \includegraphics[width=0.49\linewidth,trim={10mm 35mm 10mm 30mm},clip]{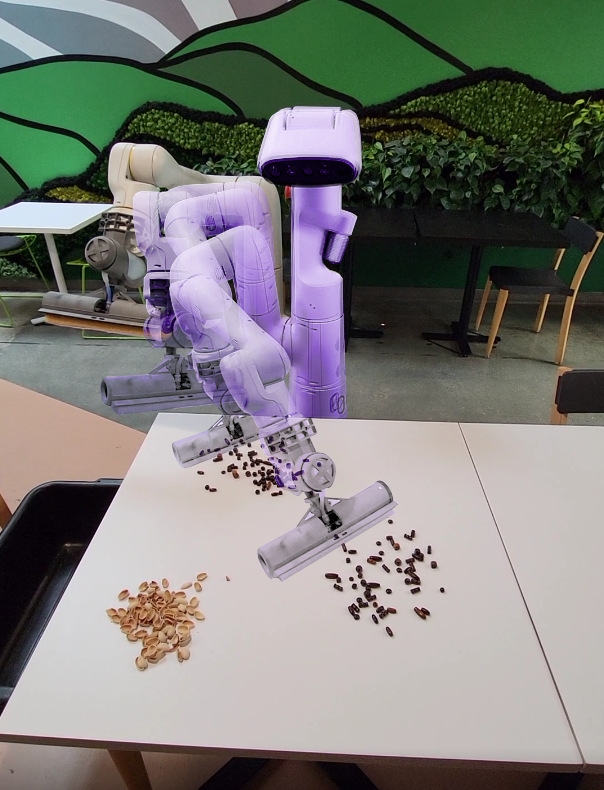}
        \includegraphics[width=0.49\linewidth,trim={10mm 80mm 10mm 74mm},clip]{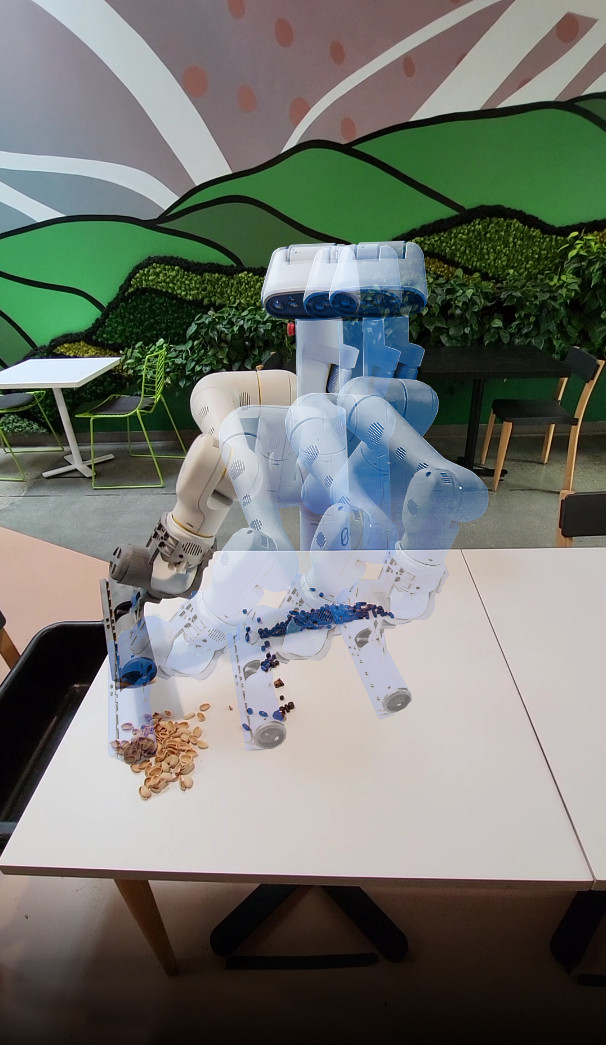}
    \caption{We present a framework to autonomously clean tables with a mobile manipulator. Our proposed approach combines reinforcement learning to select the best wiping strategy and trajectory optimization to safely execute the wiping actions. We validate our approach on the multipurpose assistive robot from \href{www.everydayrobots.com}{Everyday Robots}.}
    \vspace{-6mm}
    \label{fig:main_figure}
\end{figure}

\begin{figure*}[!t]%
        \centering
        \includegraphics[width=0.999\textwidth]{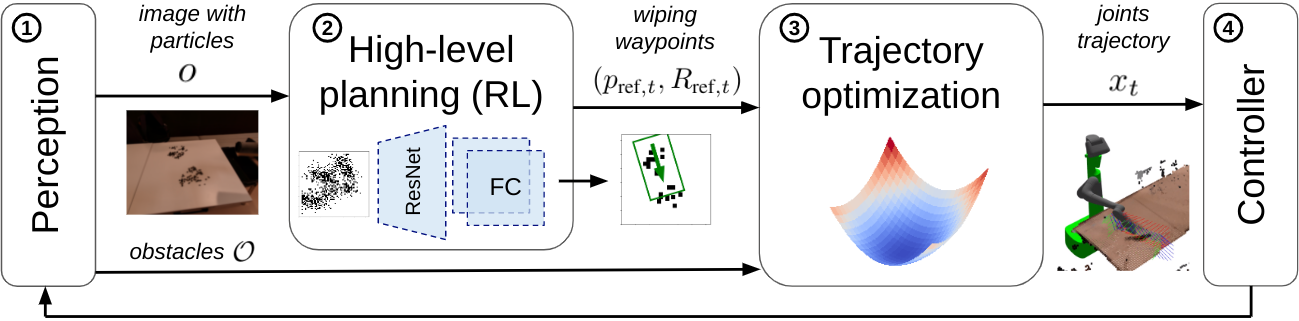}
        \begin{minipage}[t]{0.999\textwidth}
        \vspace{-1mm}
    \caption{System overview. (1) A perception module processes sensory inputs (camera images and LiDAR pointclouds) and senses obstacles, crumbs, and spills. (2) A high-level planning module selects wiping waypoints on the table. (3) A whole-body trajectory optimization module computes joint angles to perform the wipe while satisfying constraints. (4) An admittance controller executes the planned trajectory.}
    \vspace{-6mm}
    \label{fig:system_overview}
        \end{minipage}
\end{figure*}

\section{Related work}\label{sec:related_work}

\textbf{Reinforcement learning} allows tackling complex high-dimensional planning problems with stochastic multimodal dynamics that would be difficult to solve in real-time with model-based techniques  \cite{Hanson2007,ksendal2007,Theodorou2012,LewSA2022}. 
The success of RL in complex robotic tasks hinges on appropriately selecting the observation and action spaces to simplify learning \cite{Martin2019,Lee2020,Xia2021,Wu2022}. Indeed, end-to-end training is computationally costly and requires expensive data collection \cite{Kalashnikov2018,James2019,Rao2020}. 
Previous work demonstrated that decomposing complex problems by planning high-level waypoints with RL and  generating motion plans with model-based approaches improves performance \cite{Bansal2020,Xia2021}. We use a similar decomposition in this work. However, while previous approaches were applied to solve navigation  or manipulation tasks independently, we demonstrate that RL can be combined with trajectory optimization and admittance control to simultaneously move the base and arm of a mobile manipulator while avoiding obstacles to solve a complex task such as table wiping.

\textbf{Trajectory optimization} 
allows computing dynamically-feasible trajectories that guarantee reliable and safe execution, e.g., by accounting for obstacle avoidance constraints. 
Multiple works have demonstrated real-time trajectory optimization algorithms for mobile manipulators, e.g., for a ball-balancing manipulator  \cite{Minniti2019,Minniti2021} and a legged robot equipped with a robotic arm \cite{Bellicoso2019,Sleiman2021,Chiu2022}. 
In this work, we demonstrate real-time whole-body collision-free trajectory optimization %
on the mobile manipulator with an arm with seven degrees of freedom shown in Figure \ref{fig:main_figure}. The robot base has a nonholonomic constraint that makes real-time trajectory optimization challenging. 
Navigating in a cluttered environment such as a kitchen requires real-time collision avoidance. As in \cite{Schulman2014,Minniti2019,Chiu2022}, we enforce collision avoidance constraints in the formulation. 
We assume that all perceived obstacles are given as polyhedrons, as is common in the literature \cite{Majumdar2017,tordesillas2022,Marcucci2022}. 
The optimization problem is solved using the differentiable shooting-sequential-quadratic-programming (ShootingSQP) method presented in \cite{Singh2022}.

\textbf{Table wiping} with a multipurpose mobile manipulator is challenging, as this task requires simultaneously reasoning about the cleanliness state of the table, planning optimal wiping actions, and acting accordingly \cite{Kim2019}. 
Table wiping approaches can be divided into three categories. 
First, classical methods detect spills and subsequently apply pre-defined wiping patterns to clean them \cite{Yin2020}. These methods work well but are suboptimal as they do not explicitly reason about the time evolution of the spills at planning time. 
A second class of methods uses analytical or learned transition models for the cleanliness state of the table and subsequently apply classical planning methods to solve these problems \cite{Hess2011,Leidner2016,Elliott2018}.  
These methods were applied to manipulation problems too \cite{Finn2017}. 
However, \cite{Hess2011,Elliott2018} only consider dirt cleaning tasks, 
and learning transition dynamics accounting for absorption and the sticking behavior of certain liquids (e.g., honey) may be challenging. 
A third class of imitation learning methods uses demonstration data to learn a cleaning policy \cite{Kim2018,Liu2018,Gams2016,Elliott2017,Cauli2018,Shridhar2021}. These methods are successful since wiping primitives are easier to learn than visual transition models.

Our proposed approach does not require training data from the system. Instead, we train an RL policy that selects high-level wiping waypoints entirely in simulation. The key consists of describing crumbs and spill dynamics with a stochastic differential equation (SDE) \cite{Applebaum2009}. In contrast to existing learned \cite{Terry2020} models and the material point method \cite{Jiang2016}, our SDE model does not require training data from the system that may be expensive to collect, is efficient to simulate, allows modeling dry objects, sticking and absorption behavior. We then directly deploy the learned wiping policy to hardware. Wipes are executed using trajectory optimization which guarantees reliable and safe execution.

\section{Combining RL and trajectory optimization}\label{sec:system_overview}
We consider two table wiping tasks:  \textbf{gathering crumbs} and \textbf{cleaning spills}. A robot equipped with a wiper cannot immediately capture crumbs and clean dirt particles. Instead, one may first gather the crumbs together before using a different method to remove them (e.g., with a vacuum cleaner \cite{Hess2011} or by pushing them into a bin, see Section \ref{sec:results}). Thus, we formulate the objective of the crumbs-gathering task as moving all crumbs particles to the center of the table. For spills-cleaning, the objective is wiping all spill particles.  

The problem of table wiping can be formulated as a POMDP from image inputs to control signals to send to the robot actuators. In this work, we decompose the problem and propose a framework (see Figure \ref{fig:system_overview}) consisting of four steps:
\begin{enumerate}[leftmargin=6mm]
    \item \textbf{Perception}: 
    The system processes LiDAR pointclouds and camera depth and color images and returns bounding boxes for obstacles $\mathcal{O}$ (see Figure \ref{fig:robot_kinematics}) and an image mask $o$ for spills and crumbs on the table (see Section \ref{sec:RL}). 
    \item \textbf{High-level planning with RL} (Section \ref{sec:RL}): The reinforcement learning (RL) policy takes the input image with crumbs or spills on the table and returns high-level wiping waypoints, corresponding to desired start and end wiping poses of the end-effector on the table. This policy is entirely trained in simulation using the stochastic model of spill and crumbs dynamics and is directly deployed on the system without the need to gather demonstration data from the system.
    \item[3-4)] \textbf{Trajectory optimization and control} (Section \ref{sec:traj_opt}): To execute the planned wiping actions, we compute whole-body joint trajectories using trajectory optimization. This approach guarantees the satisfaction of constraints such as avoiding self-collisions and nearby obstacles such as chairs.  We track the whole-body trajectory using admittance control, which enables fine tracking while wiping with a normal force satisfying hardware requirements.
\end{enumerate}
As discussed in Sections \ref{sec:introduction} and \ref{sec:related_work}, this task decomposition leverages the strengths of reinforcement learning while enforcing constraints satisfaction with trajectory optimization. We describe these two components in the next two sections.

\section{Planning wipes with reinforcement learning}\label{sec:RL}
\subsection{Simulating spill and crumbs dynamics}
Central to our approach is a model describing the evolution of the cleanliness state of the table. It has four key features:
\begin{itemize}[leftmargin=5.5mm]
    \item It is able to describe both dry objects %
    pushed by the wiper and liquids %
    absorbed during wiping.
    \item It can capture multiple disjoint spills. 
    \item It captures the stochasticity of state transitions.
    \item It can be efficiently simulated.
\end{itemize}
We describe the cleanliness state of the table with the variable $s=(s^x,s^y,s^z)$, where $(s^x,s^y)\in\R^2$ denotes a location on the table and $s^z\in\{0,1\}$ denotes whether the corresponding crumbs or spill particles are on the table ($s^z=0$) or were wiped and are on the wiper ($s^z= 1$). 
The state at time $t$ is characterized by a measure $\mu_t$ over $\R^2 \times \{0,1\}$. %
We denote by $W_t^a\subset\R^2$ the surface of the table covered by the wiper of orientation $\theta_t$ at time $t$, as a function of the wiping action $a$ %
(see Section \ref{sec:RL:obs_action}).  
From time $t_i$, each %
particle evolves in time according to the SDE%
\begin{align}\label{eq:sde}
\dd s_t = \ &\AverageSmallMatrix{b_1(s_t,a,t)\\b_2(s_t,a,t)\\0}\dt + \AverageSmallMatrix{\alpha\sigma_1(s_t,a,t)\\\alpha\sigma_2(s_t,a,t)\\0}\dd B_t + \AverageSmallMatrix{0\\0\\h(s_t,a,t)}\dd P^\lambda_t, 
\end{align}
\\[-3mm]
where $t\in[t_i,t_i+T_i]$, 
$s_{t_i}\sim\mu_{t_i}$, and
\begin{itemize}[leftmargin=5.5mm]
    \item $B_t$ is a standard $2$-dimensional Brownian motion modeling the stochastic evolution of spill and crumbs particles,
    \item $P_t^\lambda$ is a standard Poisson process of intensity $\lambda>0$, modeling spill particles absorbed by the wiping tool, 
    \item The coefficients $b,\sigma$, and $h$ are defined by
\begin{align*}
&\hspace{-4mm}
    b(s_t,a,t)\hspace{-3.5mm}&&=\bm{1}\left\{\AverageSmallMatrix{s_t^x\\s_t^y}\in W_t^a\cap\mathcal{T}, \,  s_t^z=0\right\}v
\AverageSmallMatrix{
\cos(\theta_t) \\ \sin(\theta_t)
},
\\
&\hspace{-4mm}
    \sigma(s_t,a,t)\hspace{-3.5mm}&&=\bm{1}\left\{\AverageSmallMatrix{s_t^x\\s_t^y}\in W_t^a\cap\mathcal{T}, \,  s_t^z=0\right\}
v
\AverageSmallMatrix{
\cos(\theta_t)\, {-}\sin(\theta_t)\\
\sin(\theta_t)\ \ \, \cos(\theta_t)
},
\\
&\hspace{-4mm}
    h(s_t,a,t)\hspace{-3.5mm}&&=\bm{1}\left\{\AverageSmallMatrix{s_t^x\\s_t^y}\in W_t^a\cap\mathcal{T}, \,  s_t^z=0\right\},
\end{align*}
where $\mathcal{T}=[0,\bar{w}]\times[0,\bar{h}]$ is the table area,  $\alpha>0$ is a diffusion coefficient, and $\bm{1}(\cdot)$ is the indicator function.
\end{itemize}
The SDE \eqref{eq:sde} implies that the total mass of dirty particles is conserved over time. These particles either move on the table ($s^z=0$) or are cleaned and move onto the wiper ($s^z=1$). 
Only particles that are on the table ($s_t^z=0$) and are in contact with the wiper ($(s_t^x,s_t^y)\in W_t^a\cap\mathcal{T}$) can move.

\begin{figure}[!t]%
        \centering
        \includegraphics[width=1\linewidth,trim={0mm 0mm 0mm 0mm},clip]{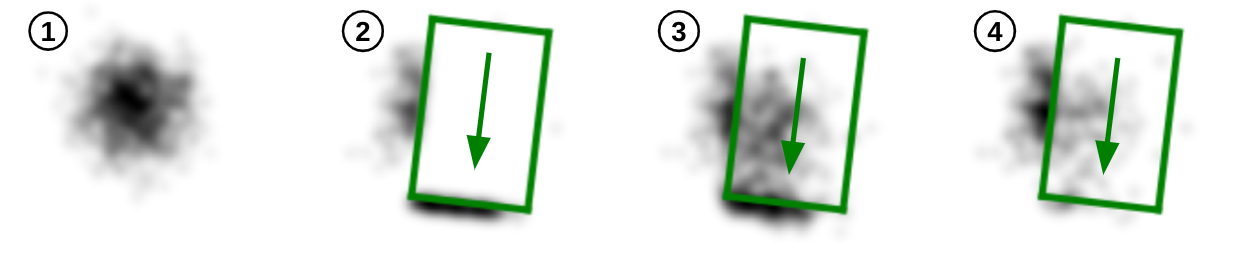}
    \caption{Spills and crumbs wiping simulator using the SDE in \eqref{eq:sde} with dirty particles in black and wiped region in green. (1) Initial state. (2)  With $(\lambda,\alpha)=(0,0)$, we simulate crumbs that are pushed by the wiper. (3) With $(\lambda,\alpha)=(0,0.05)$, we simulate a spill that is pushed on a table and sticks below the wiper. (4) With $(\lambda,\alpha)=(5,0.05)$, we simulate a spill that is partially absorbed by the wiper.}
    \label{fig:simulator}
    \vspace{-6mm}
\end{figure}

We present simulation results in Figure \ref{fig:simulator} for a single wiping action. To better represent the density of particles, we convolve the visualized particle representation with a Gaussian filter. For $\lambda\approx 0$, crumbs and dirt pushing behavior is simulated with no particles entering the wiper. By increasing $\lambda$, particles are cleaned and absorbed by the wiper. 

\subsection{Observation and action spaces}\label{sec:RL:obs_action}
\textbf{Observations}: Since only visual observations are available in practice, we convert the cleanliness state $s_t$ to an image $o(s_t)$ with $64\times64$ pixels by setting each pixel occupied by a particle to a maximum value. This approach makes transferring to the real system straightforward: only a mask defining the spill locations is necessary to deploy our approach to the robot, thereby minimizing the simulation to real gap.

\textbf{Actions}: We plan for a sequence of wipes 
$
a_i=(p^x_i,p^y_i,\theta_i,\ell_i), 
$
where $(p^x_i,p^y_i,\theta_i)$ denotes the starting position and direction of the wipe at time $t_i$ (assume $0$ w.l.o.g.) and $\ell_i$ is the wipe length. The action space is defined as $\A=[0,\bar{w}]\times[0,\bar{h}]\times[0,2\pi)\times[0,L]$, where $L=\min(\bar{w},\bar{h})$. At time $t$, the wiper covers the surface $W_t^a\subset\R^2$ of the table parameterized by $w_t^a=(w^x,w^y,w^\theta)_t^a$. 
The wiper moves at a constant speed $v$ according to 
$w_t^a=(1-t)(p^x_i,p^y_i,\theta_i) + (p^x_i+tv\cos(\theta_i),p^y_i+tv\sin(\theta_i),\theta_i)$ 
where $t\in[0, T_i]$ with $T_i=\frac{\ell_i}{v}$. This four-dimensional action space makes planning tractable, %
albeit one could also use pre-defined \cite{Yin2020} or learned \cite{Kim2018,Elliott2017} wiping patterns instead. We then convert $a_i$ to wiping poses  $(p^{\textrm{ref}}_t, R^{\textrm{ref}}_t)$ by  aligning the wipe with the pose of the table, see Section \ref{sec:results} for details.

\subsection{Wiping objectives and constraints}
We define objectives for the tasks defined in Section \ref{sec:system_overview}.

\textbf{Gathering crumbs}: A simple reward for this problem is
\begin{equation}\label{eq:reward:gathering}
R(s,a) =
-%
    \left\|(s^x,s^y)-\left(\frac{\bar{w}}{2},\frac{\bar{h}}{2}\right)\right\|_2
\end{equation}
to penalize the spread of dirty particles to the table center. 

\textbf{Cleaning spills} can be described as absorbing all spills with the wiper, i.e., maximizing particles with $s^z= 1$. %
An objective for this task could be maximizing $\E\left[s^z\right]$. In practice, we found that expressing the objective as function of pixel values works better for this task since the latent state $s^z$ is not observable. Thus, we express the objective  as
\begin{equation}\label{eq:reward:wiping}
R(s,a) = %
\sum_{ij} (o_{ij}(s_{t+1})-o_{ij}(s_t))
\end{equation}

We do not minimize the wiping lengths $\ell_i$, since they do not exactly reflect the wiping duration that depends on the full kinematics of the robot and obstacles in the environment. 

Wiping requires keeping crumbs and spill particles on the table. Due to the stochasticity in \eqref{eq:sde} and the initial particles distribution $\mu_0$, we express this requirement with a joint chance constraint \cite{LewEtAl2022_rl}  
$\mathbb{P}\left((s^x_t,s^y_t) \in\mathcal{T}\, \text{for all}\,  t\in[0,T]\right)\geq 1- \delta$ for some probability threshold $\delta\in(0,1)$. %
Since particles are immobile when not in contact with the wiper, %
this constraint is equivalent to the joint chance constraint $\mathbb{P}\left((s^x_{t_i},s^y_{t_i})\in\mathcal{T}\, \text{for all}\   i=1,\dots,N\right)\geq 1-\delta$. Using Boole's inequality, a conservative reformulation is given by %
the expectation constraints  $\E\left[\bm{1}\{s_{t_i}\notin\text{Table}\}\right]\leq\frac{\delta}{N}$ for all $i$.

\subsection{Solving the problem via reinforcement learning}
By considering a sequence of $N$ wipes and summing the objectives in \eqref{eq:reward:gathering} and \eqref{eq:reward:wiping}, 
we obtain the constrained Markov decision process (CMDP) \cite{Altman1999}
\begin{subequations}
\begin{align*}
\sup_{\substack{a_i\in\A \\ i=1,\dots,N}}
\E\left[\sum_{i=1}^N R(s_{t_i}, a_i)\right]
\, 
\textrm{s.t.}
\  
\eqref{eq:sde}, 
\ 
\E\left[\bm{1}\{s_{t_i}\notin\mathcal{T}\}\right]\leq\frac{\delta}{N}.
\end{align*}
\end{subequations}
where each action $a_i$ only depends on visual observations $\{o(s_{t_j}), j\leq i\}$. This MDP problem structure assumes that visual observations $o(s)$ are expressive enough to reconstruct the state $s$. Without this assumption, the problem above is a constrained partially-observable MDP (POMDP). %

\begin{figure}[!t]
        \centering
        \includegraphics[width=1\linewidth,trim={0mm 0mm 0mm 0mm},clip]{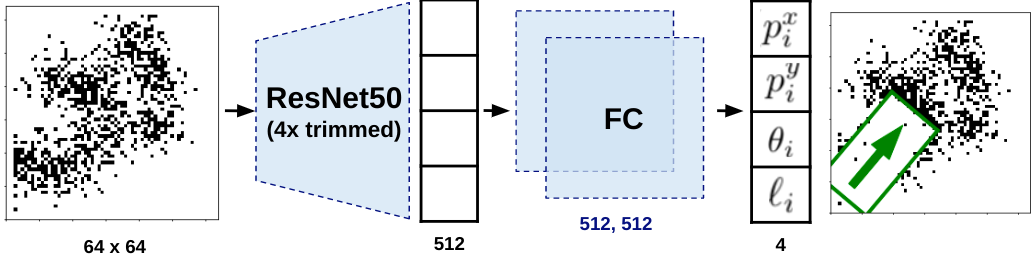}
    \caption{SAC actor network architecture.}
    \label{fig:nn_architecture}
    \vspace{-6mm}
\end{figure}

Due to the high-dimensional visual observation space, the complexity of the stochastic dynamics in \eqref{eq:sde}, and the potentially multimodal state distribution, solving the CMDP is challenging. Since simulation is efficient and can be carried out in parallel, we solve the problem using RL. A wide range of RL approaches for CMDPs are available in the literature, %
ranging from methods using Lyapunov functions to restrict the search to feasible policies \cite{Chow2018,Chow2020}, using learned dynamics models to predict failures and yield a safe policy \cite{Thomas2021,LewEtAl2022_rl}, and safety filters that modify the actions of a policy to yield constraints satisfaction \cite{Alshiekh2018}, to name a few. We refer to \cite{Gu2022} for a recent survey. 

Since the policy can be trained in simulation, for simplicity, we opt for constraints penalization via a Lagrangian relaxation %
and solve the relaxed problem
\begin{subequations}
\begin{align*}
\sup_{\substack{a_i\in\A \\ i=1,\dots,N}}
\E\left[\sum_{i=1}^N R(s_{t_i}, a_i) - \mu
\bm{1}\{s_{t_i}\notin\mathcal{T}\}\right]
\ \ 
\textrm{s.t.}
\ \ 
\eqref{eq:sde}, 
\end{align*}
\end{subequations}
where $\mu>0$ is a penalization weight. The problem above can be solved using off-the-shelf RL algorithms.  
We solve the MDP with the soft actor-critic (SAC) method \cite{Haarnoja2018} due to its robustness and sample efficiency and provide further implementation details in Section \ref{sec:results}.

\section{Executing wipes with trajectory optimization}\label{sec:traj_opt}

\begin{figure}[!t]%
\centering
\begin{minipage}{0.6\linewidth}
\includegraphics[width=0.9\linewidth,trim={0mm 7mm 0mm 5mm},clip]{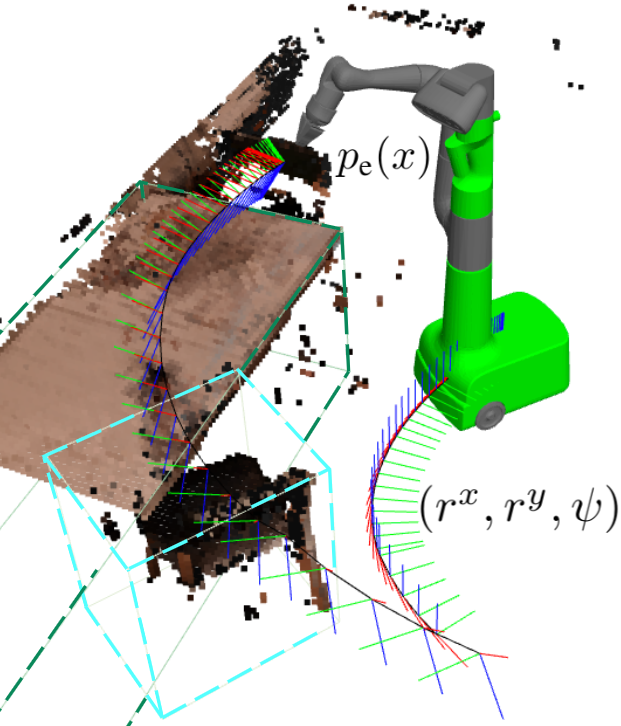}
\end{minipage}
\begin{minipage}{0.38\linewidth}
    \caption{The robot from \href{www.everydayrobots.com}{Everyday Robots} consists of a manipulator with seven joints $(q^1,\dots,q^7)$ mounted on a base with two wheels whose position and yaw is denoted by $(r^x,r^y,\psi)$. The kinematics of the base account for nonholonomic constraints. We enforce collision avoidance constraints within the trajectory optimization formulation using bounding boxes for obstacles.}
    \label{fig:robot_kinematics}
\end{minipage}
\vspace{-7mm}
\end{figure}

Next, we describe the whole-body trajectory optimization formulation we use to execute wiping actions with the mobile manipulator shown in Figure \ref{fig:robot_kinematics}. 
Since the system is passively stable, we plan for kinematically-feasible trajectories. The state of the robot is denoted by $x=(r^x,r^y,\psi,q^1,\dots,q^7)$, where $(r^x,r^y,\psi)$ denotes the position and yaw-orientation of the robot base, and $q^i$ denotes the angle of the $i$th joint of the arm. The control input $u=(u^r,u^\psi,u^1,\dots,u^7)$ corresponds to the forward and angular velocity of the base and to each $i$th joint velocity. The system follows the dynamics 
\begin{equation}\label{eq:dynamics}
(\dot{r}^x,\dot{r}^y)=(\cos(\psi),\sin(\psi))u^r,
\quad
\dot\psi=u^\psi,
\quad \dot{q}^i=u^i,
\end{equation}
where $i=1,\dots,7$. The position and rotation matrix of the end-effector are denoted by $p_{\textrm{e}}(x)$ and $R_{\textrm{e}}(x)$, respectively. They are defined by the forward kinematics of the system and are computed as a chain of homogeneous transformation matrices $T_i(x)\in SE(3)$, %
computed using the product of exponentials formula ~\cite{lynch2017modern}. %
Given a wiping trajectory $T^{\textrm{ref}}_t=(p^{\textrm{ref}}_t,R^{\textrm{ref}}_t)$, we minimize the cost function %
\begin{equation}\label{eq:cost}
\inf_u 
\int_0^T 
\left(\ell^u(u_t)+\ell^x(x_t,t)\right)
\dt,
\end{equation}
where $T>0$ is a time-horizon, $\ell^u(u_t)=\|u_t\|^2$ penalizes the control effort, and 
$\ell^x(x_t,t)=\|p_\textrm{e}(x_t)-p^{\textrm{ref}}_t\|^2+\|I-R_\textrm{e}(x_t)^\top R^{\textrm{ref}}_t\|$ minimizes the end-effector pose tracking error.

We consider two types of obstacle avoidance constraints. First, similarly to \cite{Chiu2022}, self-collisions are avoided by covering the robot with spheres $(p_i(x),r_i)$ and enforcing 
\begin{equation}\label{eq:self_avoid}
\|p_i(x)-p_j(x)\|\geq r_i+r_j
\end{equation}
for all pairs of indices $i\neq j$ that correspond to potential self-intersections given the kinematics of the robot. 

Second, we enforce collision avoidance constraints with polytopic obstacles $\Obs$ detected by the perception stack. 
Each obstacle is expressed by $q\in\N$ linear inequalities $
\Obs=\{p\in\R^3:A(p-c)\leq b\}
$ with $A\in \R^{q\times 3}$, $c\in\R^3$, and $b\in\R^q$, where each $j$th row $A_{j\cdot}$ of $A$ is unit-norm. In practice, we use bounding boxes to enclose obstacles and represent them as polytopes, see Figure \ref{fig:robot_kinematics}.  
Multiple approaches to enforce collision avoidance constraints with such obstacles exist in the literature, such as decomposing the feasible workspace into convex regions \cite{Marcucci2022}, enforcing linearized signed-distance constraints \cite{Majumdar2017,LewBonalliEtAl2020}, or solving a convex program \cite{Tracy2022}. We propose a different approach in this work. 
For any point $p$, we first compute the index $j^\star(p)=\mathop{\arg\min}_{j=1,\dots,q} \{\gamma_j: \gamma_j\geq 0, 
\gamma_jA_{j\cdot}^\top (p-c) = b_j\}$ corresponding to the facet of $\mathcal{O}$ that intersects the line from $p$ to $c$. 
Then, for any sphere $(p_i(x),r_i)$ covering the robot, we enforce the corresponding $r_i$-padded halfplane constraint
\vspace{-6mm}

\begin{equation}\label{eq:obs_avoid}
A_{j_i^\star\cdot}^\top (p_i(x)-c)\geq b_{j^\star}+r_i,
\end{equation}
where $j_i^\star = j^\star(p_i(x))$. This constraint %
guarantees collision avoidance since $\|A_{j\cdot}\|{=}1$ and $\Obs$ is convex. It is differentiable almost everywhere and easy to implement. 
Using \eqref{eq:dynamics}-\eqref{eq:obs_avoid}, for an initial state $x^0$, we obtain the optimal control problem:
\begin{subequations}\label{eq:OCP}
\vspace{-1mm}
\begin{align}
\textbf{OCP}: \ 
\inf_{u\in\U}\ \ 
&\int_0^T \ell(x_t,u_t,t)\dt 
\quad \textrm{s.t.} \ \ x_0=x^0, 
\\
&\dot{x}_t=f(x_t,u_t),  
\ 
c(x_t)\geq 0, \  t\in[0,T].
\end{align}
\end{subequations}
All terms in \textbf{OCP} are twice differentiable almost everywhere. We discretize \textbf{OCP} with an Euler scheme, which results in an optimization problem that can be solved with any off-the-shelf solver. In this work, we leverage the sequential-quadratic-programming shooting method presented in \cite{Singh2022}: a computationally efficient method with stable convergence properties, yielding a resilient implementation on the system. 

\textbf{Feedback control}: We track the whole-body joint angles using admittance control. This approach enables fine trajectory tracking while wiping with a desired normal force of $10\textrm{N}$ that meets hardware requirements.

\section{Experimental evaluation}\label{sec:results}

\textbf{Offline RL training}: We train two wiping policies for crumbs-gathering and spills-cleaning with the same network architecture shown in Figure \ref{fig:nn_architecture}. 
For spills wiping, we terminate each episode once there are no visible dirty pixels in the input image. For crumbs-gathering, we terminate once the error in \eqref{eq:reward:gathering} is smaller than $0.02$, which corresponds to having all particles within a $15$cm radius around the table center. We implement the penalization term $\bm{1}\{s_{t_i}\notin\mathcal{T}\}$ by penalizing with a constant term if at least one particle is out of the table, which we found stabilizes training. We train for $50$k training and $50$ million environment steps with a maximum episode length $N=20$ and a batch size of $256$. 

\begin{figure}[!t]
\begin{minipage}{0.6\linewidth}
        \includegraphics[width=0.32\linewidth,trim={7mm 7mm 2mm 0mm},clip]{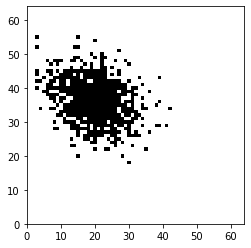}
\includegraphics[width=0.32\linewidth,trim={7mm 7mm 2mm 0mm},clip]{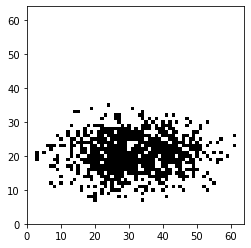}
\includegraphics[width=0.32\linewidth,trim={7mm 7mm 2mm 0mm},clip]{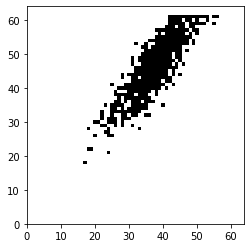}
\end{minipage}
\hspace{1mm}
\begin{minipage}{0.3\linewidth}
    \caption{Initial state distributions for RL training.}
    \label{fig:init_sim_states}
    \end{minipage}
    \vspace{-4mm}
\end{figure}

We simulate the SDE using $10^3$ particles, discretizing \eqref{eq:sde} with an Euler-Maruyama scheme with $\Delta t=0.1$s. Simulating a wipe takes only $5$ms with a Python implementation, measured on a computer with a 2.20GHz Intel Xeon CPU.  
We use $(\alpha,\lambda)\,{=}\,(10^{-2},0)$ and  $(\alpha,\lambda)\,{=}\,(10^{-2},2)$ for the crumbs- and spills-wiping environments. We randomize the initial state by sampling particles from Gaussian distributions, see Figure \ref{fig:init_sim_states}. 
We consider $1\,{\times}\,1\, \textrm{m}$ tables and a rectangular $30\,{\times}\,5 \, \textrm{cm}$ wiper $W_t^a$ moving at a constant speed $v=15\,\frac{\textrm{cm}}{s}$.

\begin{figure}[!htb]
        \centering
        \includegraphics[width=0.48\linewidth,trim={0mm 15mm 0mm 0mm},clip]{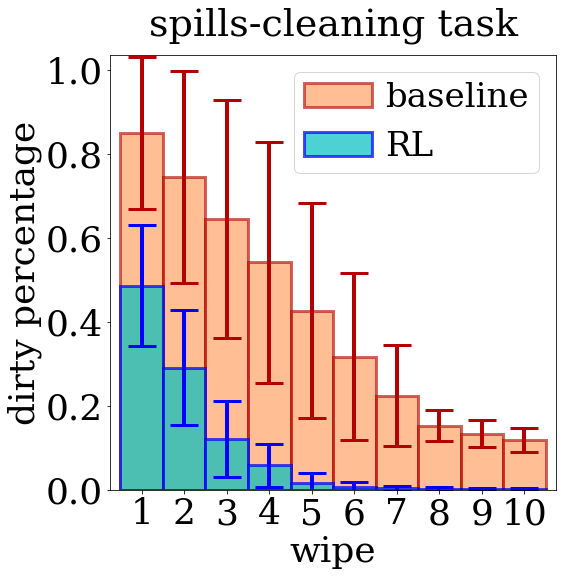}
        \includegraphics[width=0.5\linewidth,trim={0mm 15mm 0mm 0mm},clip]{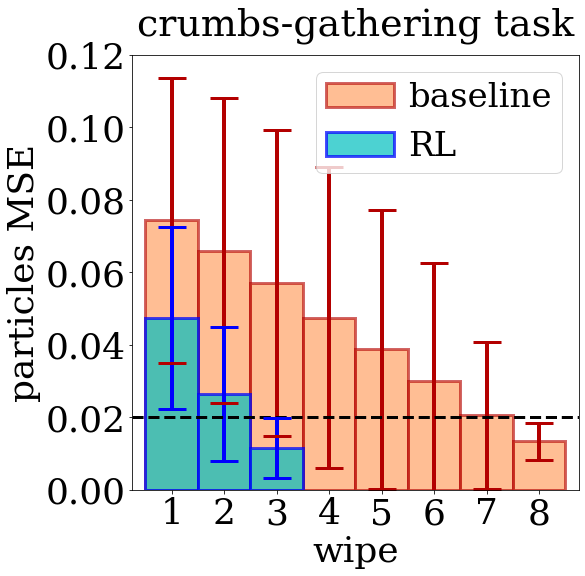}
    \caption{Comparisons between the RL wiping policy and the baseline. }
    \label{fig:RLperformance}
\end{figure}

\textbf{Simulation results}: 
We validate our RL policies in the environments with initial distributions shown in Figure \ref{fig:init_sim_states}, and compare them with a baseline that always wipes to the center of the table with an orientation rotating around the table in increments of $\pi/4$. We also compared with a method that wipes along the direction of largest covariance, which we found had worse performance due to the multimodal particle distribution after the initial wipe. We present results for aggregated $10^3$ rollouts in Figure \ref{fig:RLperformance}. Error bars correspond to $\pm 1$ standard deviation. Results demonstrate that our RL policies wipe significantly faster than the baseline. %

\begin{figure}[!htb]
\begin{minipage}[t]{1\linewidth}
        \includegraphics[width=1\linewidth]{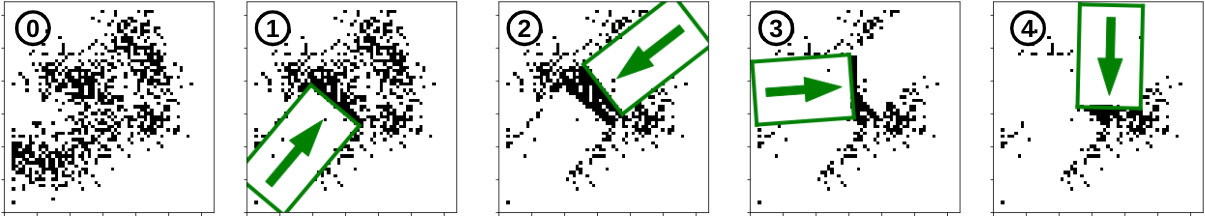}
        \includegraphics[width=0.38\linewidth]{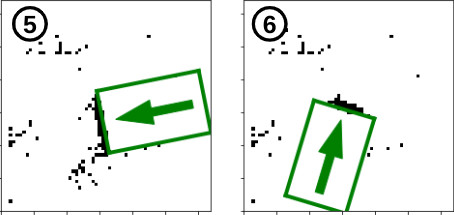}
    \hspace{1mm}
    \begin{minipage}[t]{0.58\linewidth}
    \vspace{-13mm}
    \caption{Rollout of the crumbs-gathering (top two lines) and spills-cleaning (bottom two lines) RL policies.}
    \label{fig:sim:RL}
    \end{minipage}
\end{minipage}
        \vfill
        \vspace{1mm}
        \vfill
\begin{minipage}[t]{1\linewidth}
        \includegraphics[width=1\linewidth]{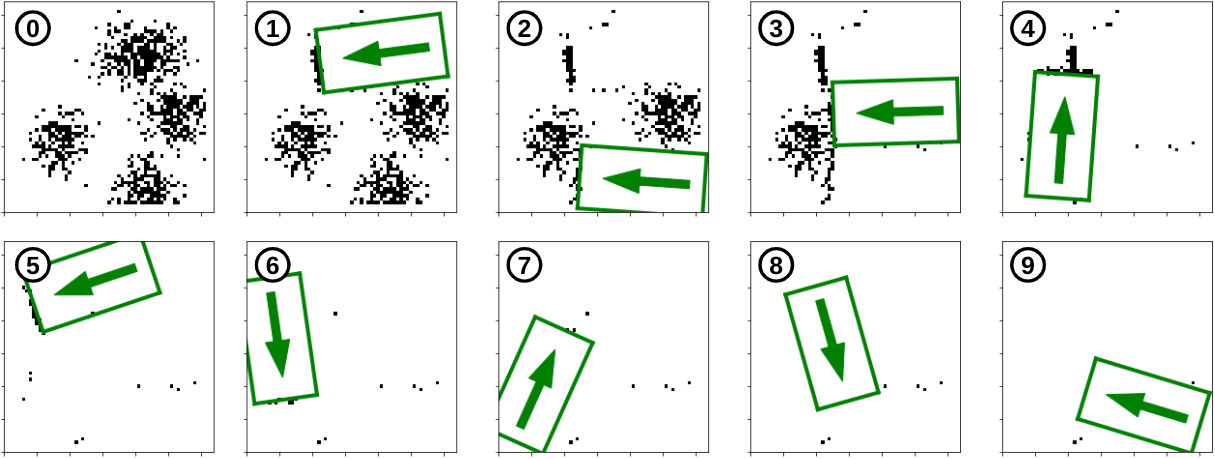}
\end{minipage}
\end{figure}

We test these two policies on environments with initial states sampled from a mixture of Gaussians representing multiple unclean table areas and show results in Figure \ref{fig:sim:RL}. We observe that despite having never encountered these states at training time, the RL policies generalize to these observations. This is mostly likely due to the inductive bias from the CNN policy architecture. Remarkably, the policy never wipes out of the table in our experiments.

\begin{figure*}[!t]%
\begin{minipage}[[t]{1\textwidth}
        \centering
        \includegraphics[width=0.195\linewidth,trim={160mm 0mm 150mm 0mm},clip]{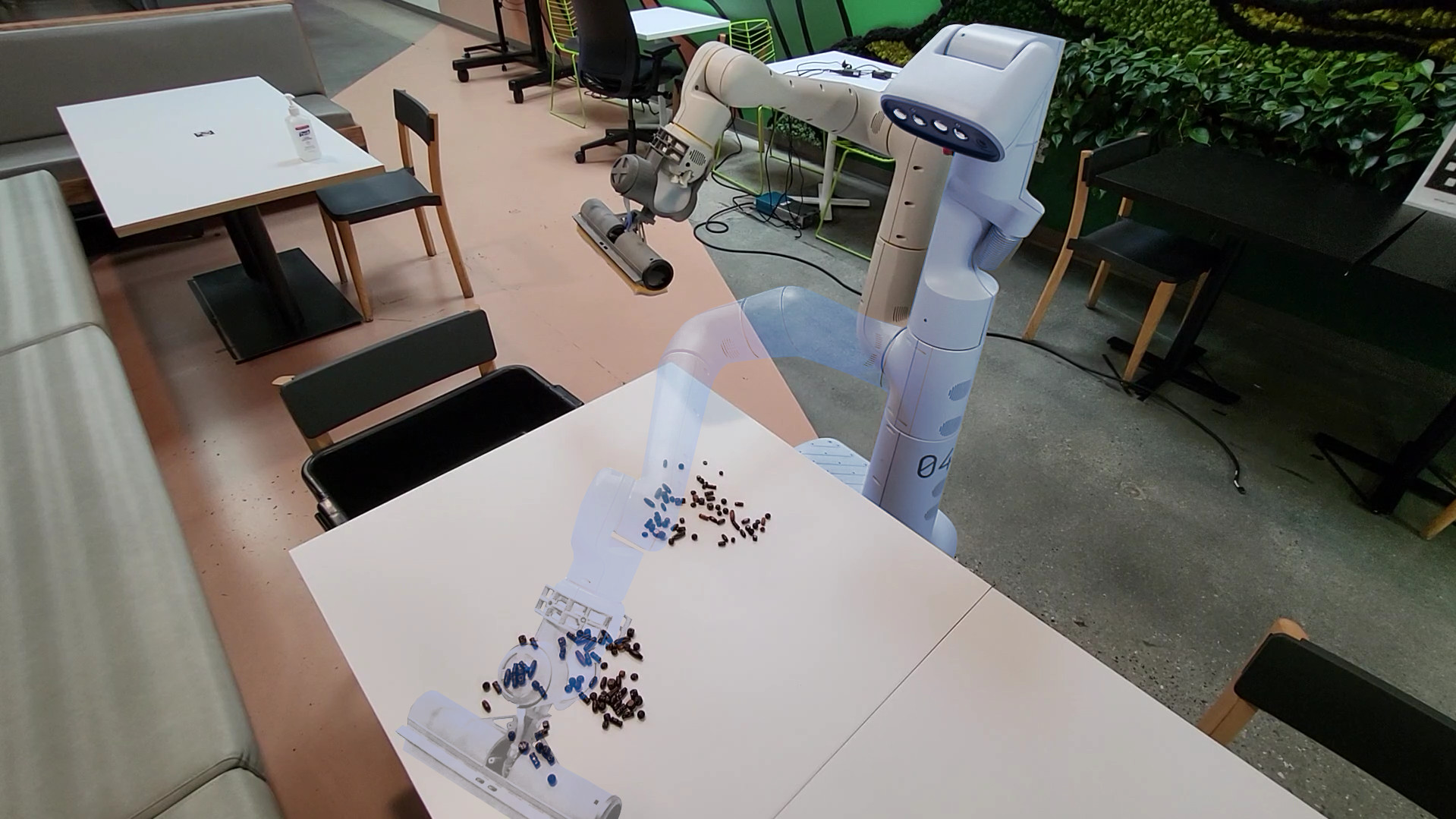}
        \includegraphics[width=0.195\linewidth,trim={160mm 0mm 150mm 0mm},clip]{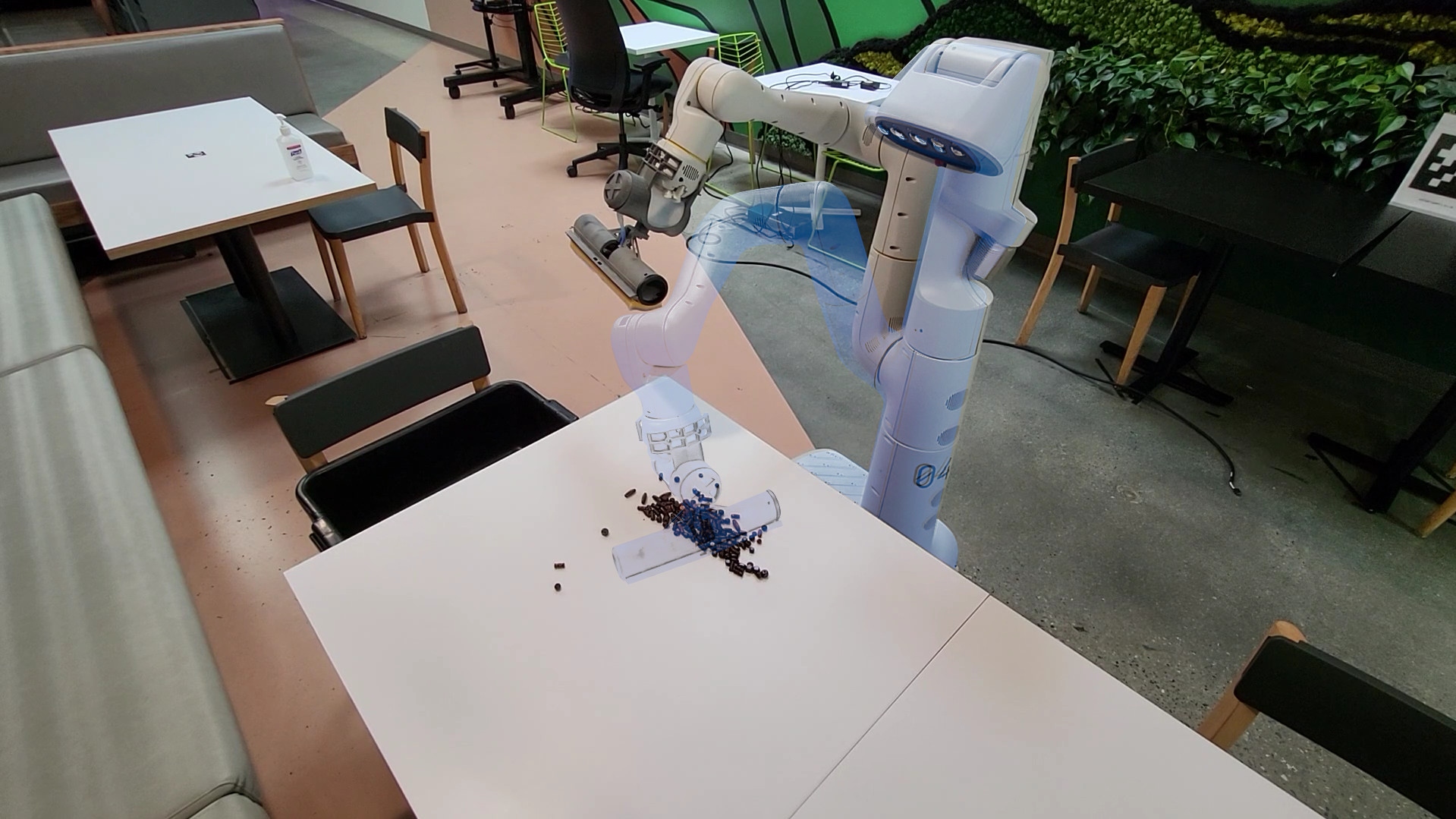}
        \includegraphics[width=0.195\linewidth,trim={160mm 0mm 150mm 0mm},clip]{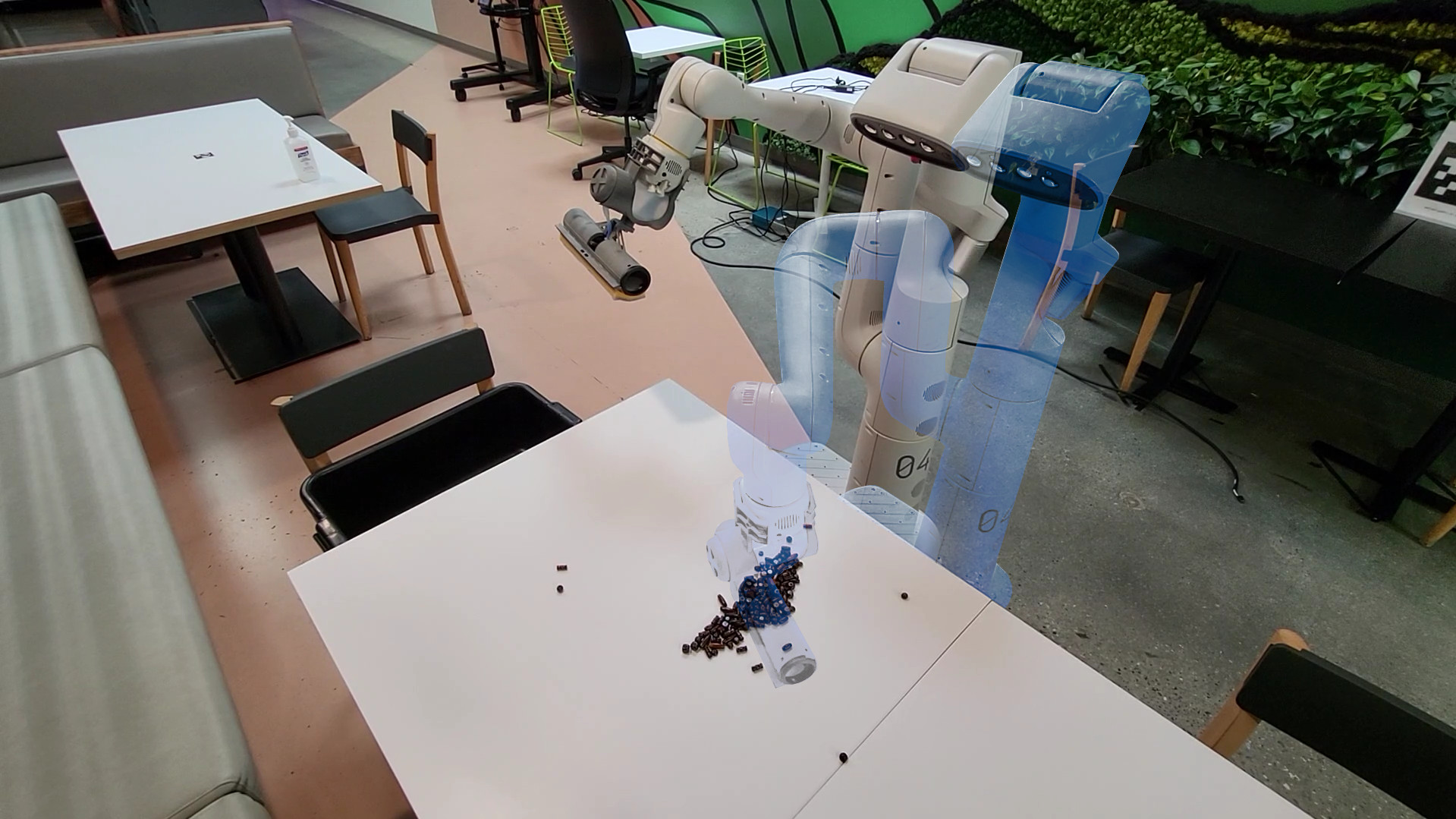}
        \includegraphics[width=0.195\linewidth,trim={160mm 0mm 150mm 0mm},clip]{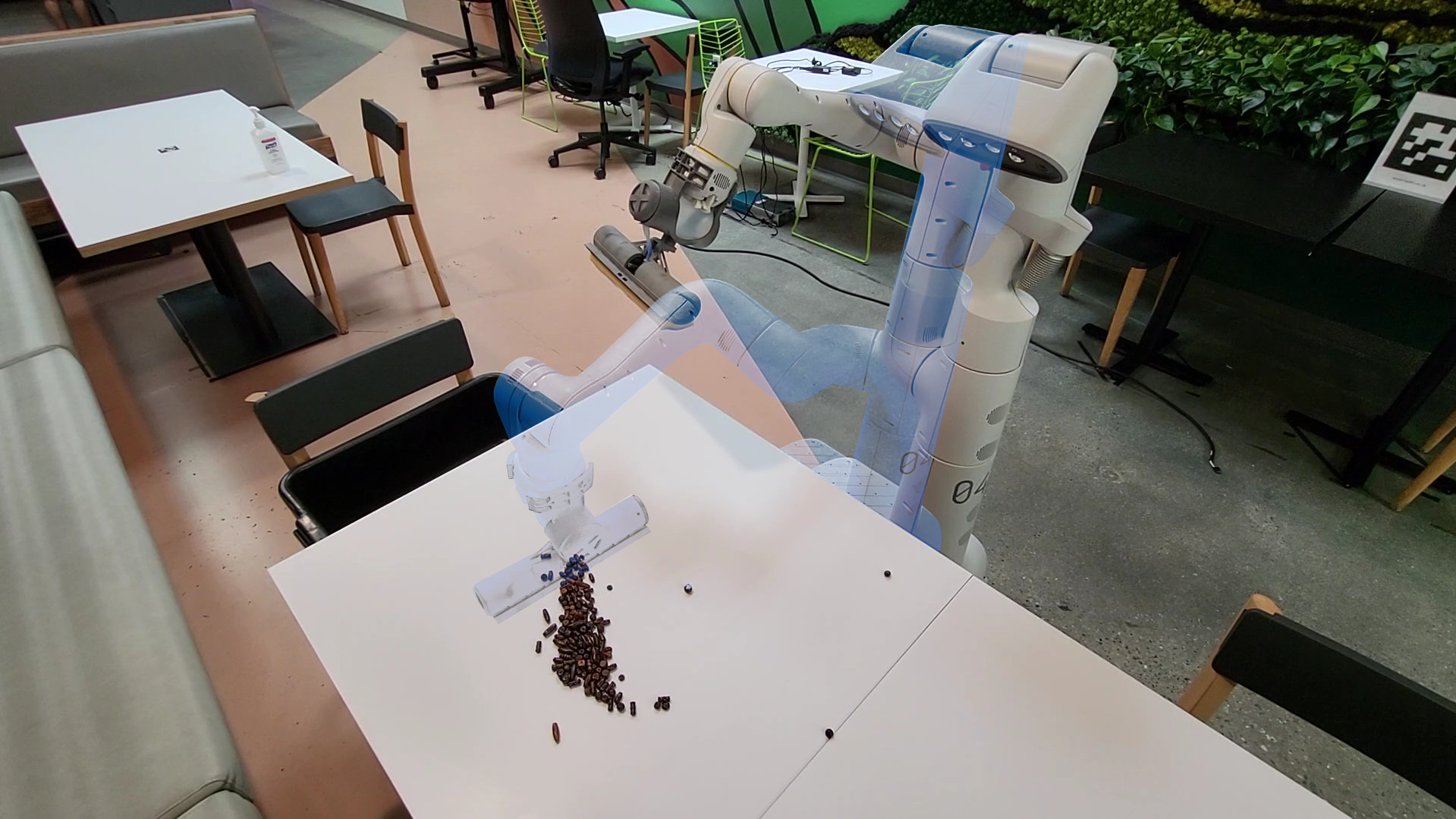}
        \includegraphics[width=0.195\linewidth,trim={160mm 0mm 150mm 0mm},clip]{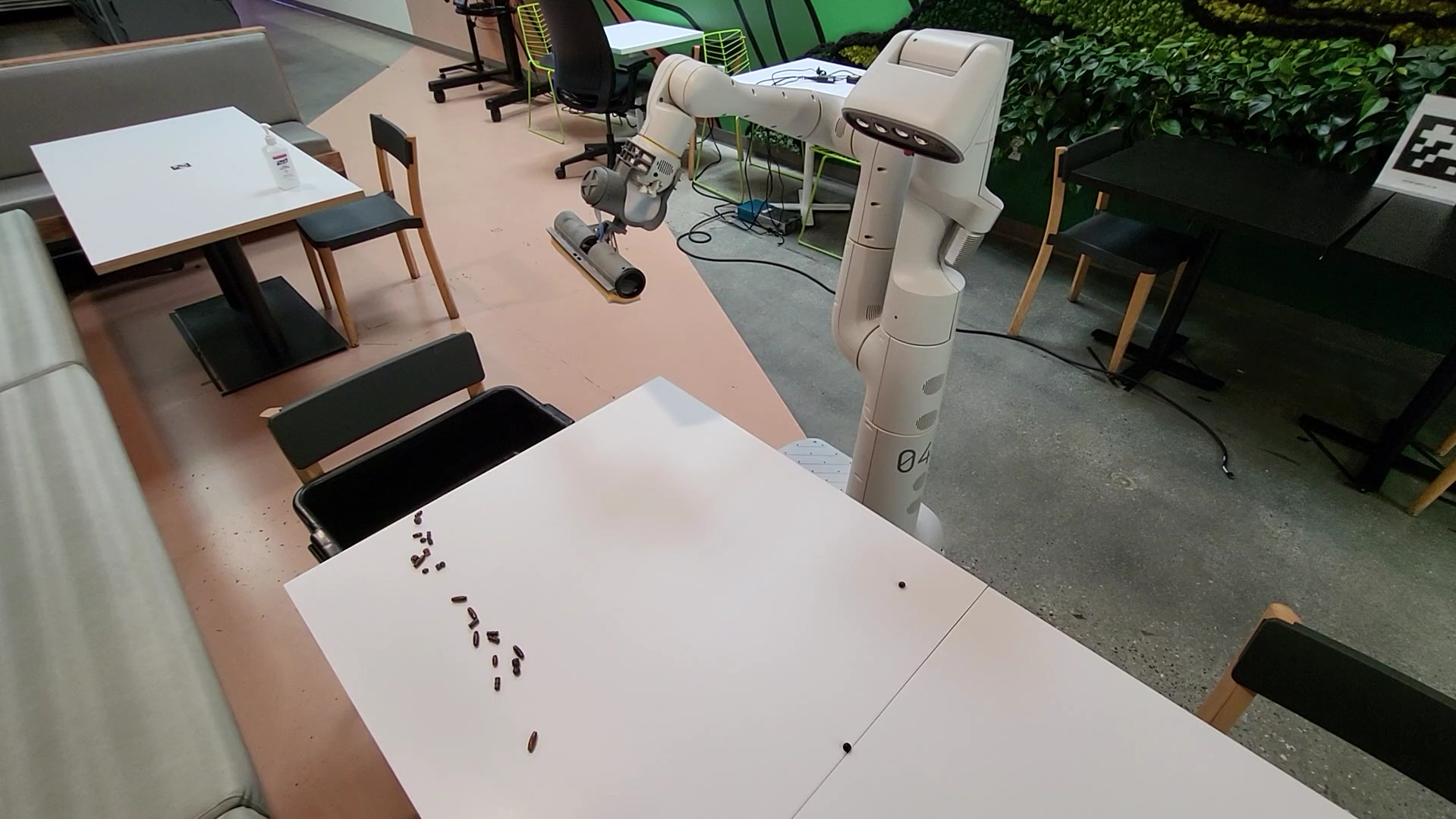}
\end{minipage}
        \vspace{1mm}
        \vfill
\begin{minipage}[t]{1\textwidth}
        \includegraphics[width=0.15\linewidth,trim={7mm 7mm 2mm 0mm},clip]{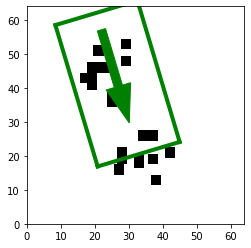}
        \includegraphics[width=0.15\linewidth,trim={0mm 0mm 0mm 0mm},clip]{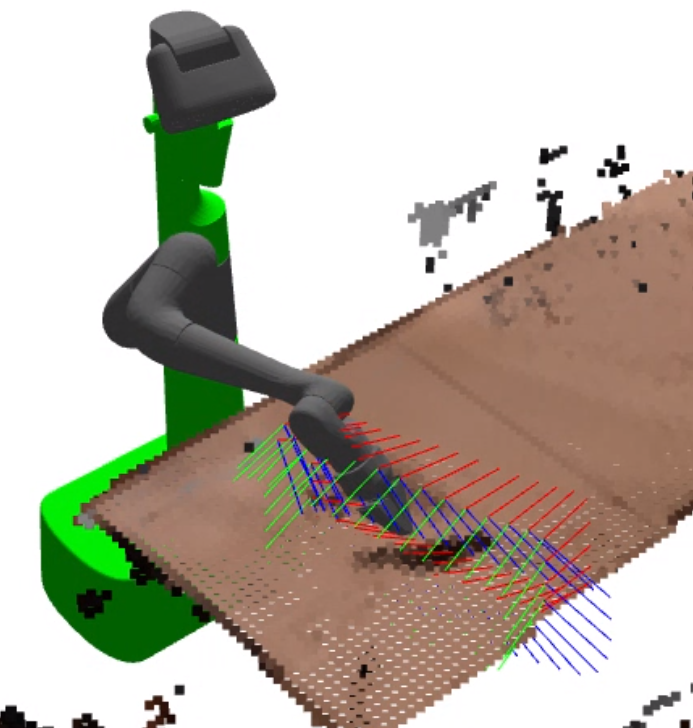}
        \hspace{15mm}
        \includegraphics[width=0.195\linewidth,trim={50mm 0mm 50mm 0mm},clip]{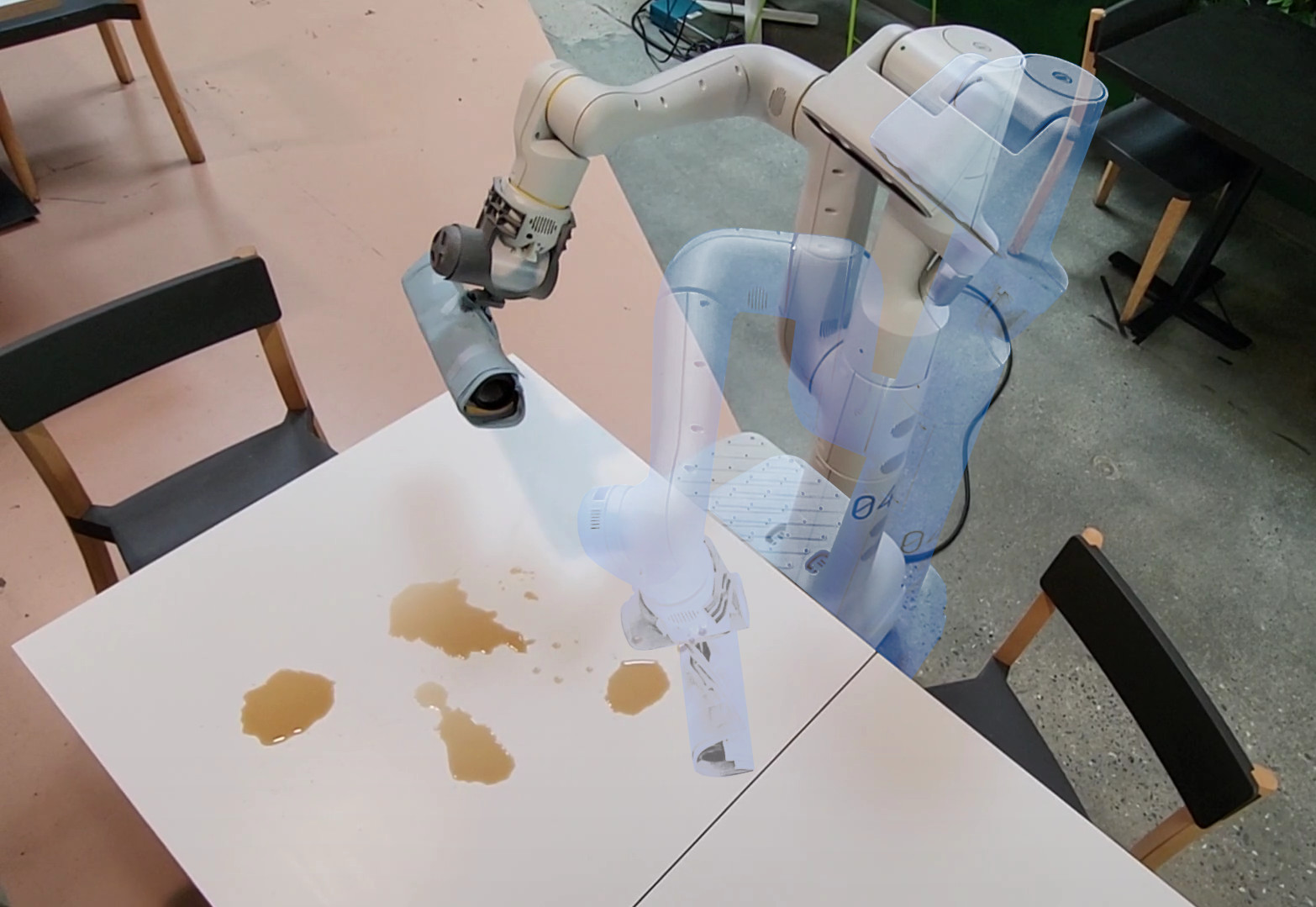}
        \includegraphics[width=0.195\linewidth,trim={50mm 0mm 50mm 0mm},clip]{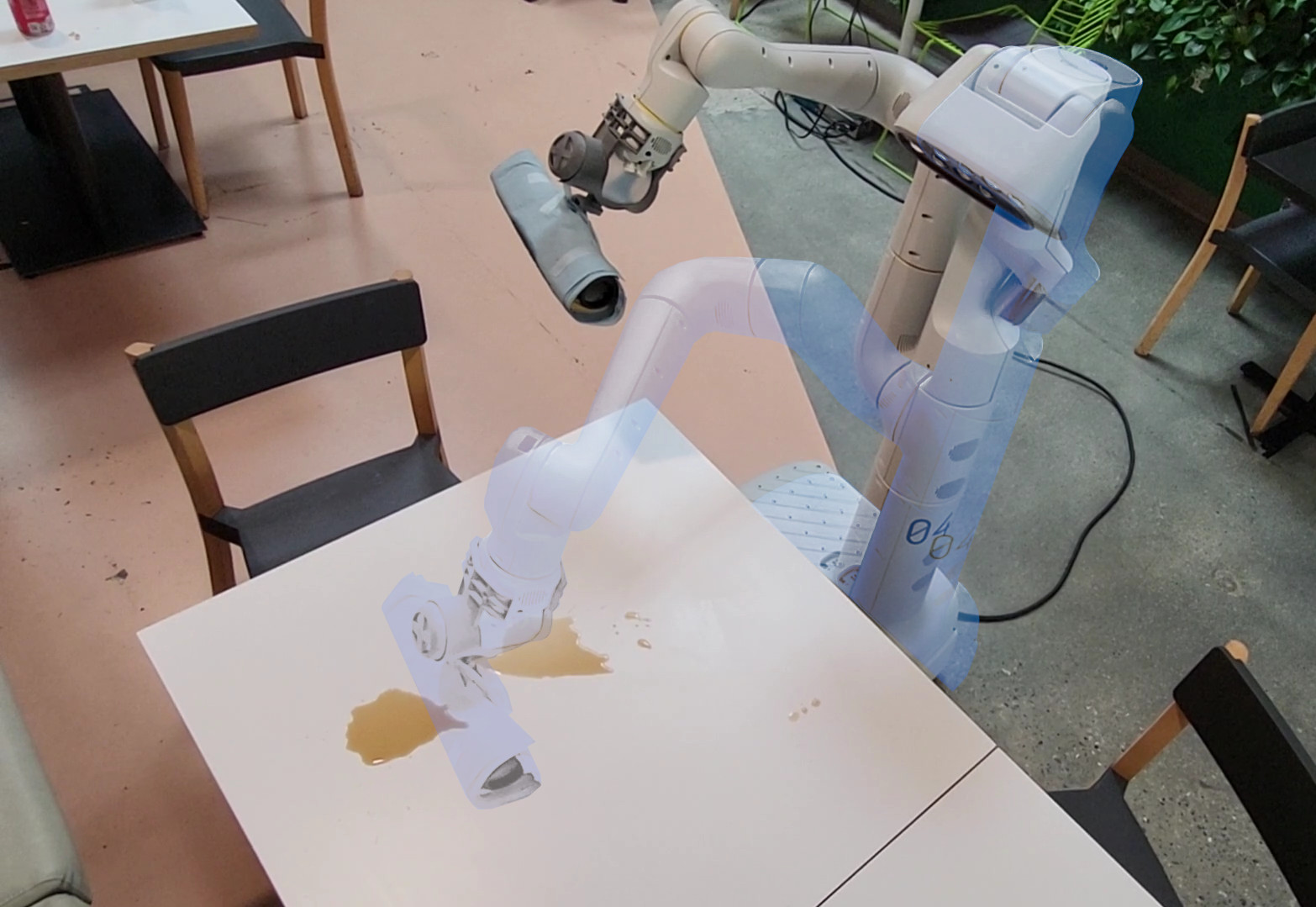}
        \includegraphics[width=0.195\linewidth,trim={50mm 0mm 50mm 0mm},clip]{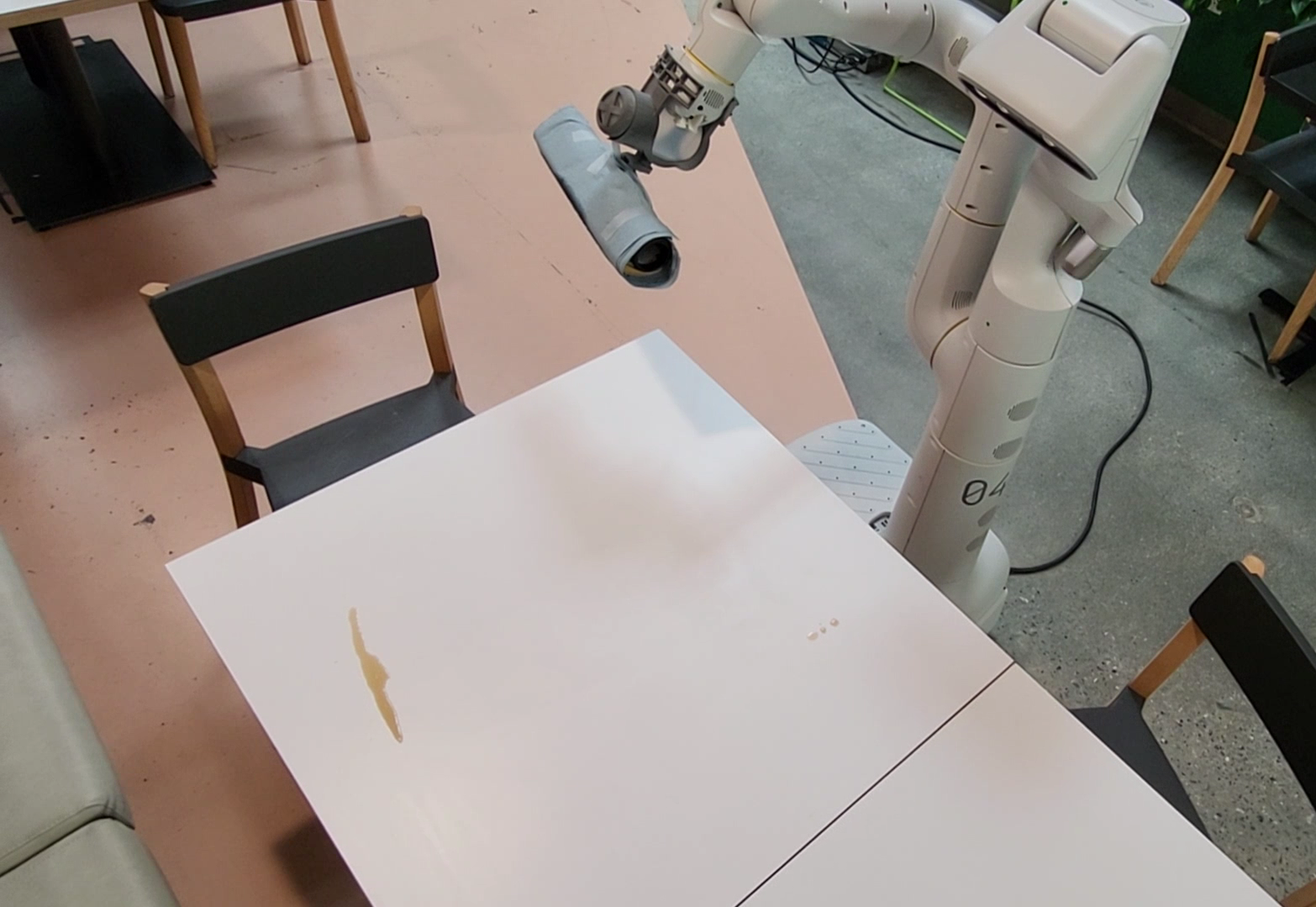}
\end{minipage}
        \vspace{-4mm}
        \begin{minipage}[t]{1\textwidth}
        \vspace{-1mm}
    \caption{Results on the assistive robot from \href{www.everydayrobots.com}{Everyday Robots} on crumbs-gathering (top and bottom left) and spills-wiping tasks (bottom).} 
    \label{fig:results:hardware}
        \end{minipage}
\end{figure*}

\textbf{Hardware setup}: We consider crumbs-gathering and spill-wiping scenarios on a table with two chairs that limit the motion of the robot. For crumbs-gathering tasks, we compute a wiping trajectory to a bin once all crumbs are closely together. For  spills-wiping, we terminate the task once no particles are detected by the perception module. 
To obtain inputs for the RL policy, we use color filtering from images to obtain a mask $o$ of resolution $64{\times} 64$ pixels with approximate locations of dirty particles on the table. 
Since this outputs a sparse representation of perceived spill and crumbs, we inflate the detected particles by two pixels before passing this input to the RL policy. Inference time of the network is $7\,\textrm{ms}$. 
Wiping actions are then converted to a desired pose trajectory $(p^{\textrm{ref}}_t, R^{\textrm{ref}}_t)$ on the table passing from the current tool pose, the wiping poses, and back to the initial tool pose. To wipe effectively, we minimize the yaw and roll orientation error but do not consider the pitch angle of the end-effector with respect to the table for trajectory optimization. Computing a whole-body collision-free trajectory takes about $500\,\textrm{ms}$ with zero-controls as the initial guess. 
Finally, the admittance controller tracks the joint trajectory while applying a $10\,\textrm{N}$ normal force with the table, meeting hardware requirements and yielding satisfactory wiping results.

\textbf{Hardware results}: 
We present results in Figure \ref{fig:results:hardware} and in the supplementary video.  
We observe that the RL policy generalizes well, thanks to the design of the observation space that minimizes the simulation to real gap. In all scenarios, the wipes planned by the RL policy point in the direction of the center of the table to avoid moving crumbs and spills out of the table. The trajectory optimizer computes whole-body trajectories that successfully avoid obstacles with chairs and the table, while operating in close proximity with obstacles.

\begin{figure}[!htb]
        \centering
    \vspace{-1mm}
        \includegraphics[width=1\linewidth,trim={0mm 0mm 0mm 0mm},clip]{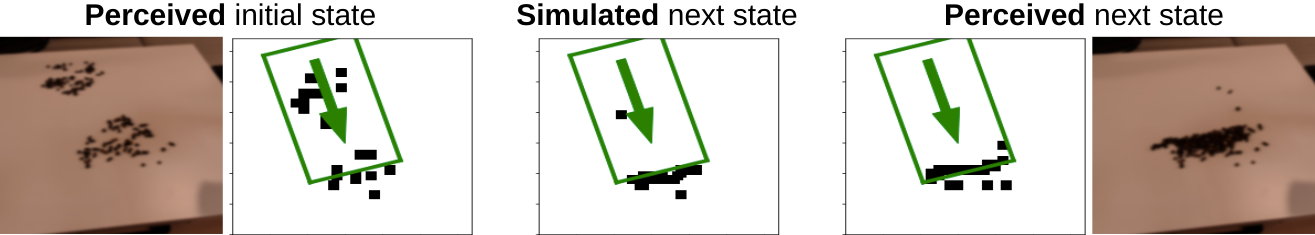}
    \caption{Initial image and planned wipe (left), prediction after wipe from simulator (middle), and final observation after wiping (right).}
    \label{fig:results:crumbs:sim_accurate}
    \vspace{-2mm}
\end{figure}

To validate our simulator, we use observations from the system to initialize the simulator and predict the evolution of perceived particles for a planned wipe. Then, we compare the result with perceived particles after executing the wipe on the system. In Figure \ref{fig:results:crumbs:sim_accurate}, we present results for the first wipe of the crumbs-gathering scenario shown in Figure \ref{fig:results:hardware}. 
We observe that predictions are reasonable and correctly predict the general trend of crumbs particles. This result explains the success of hardware experiments: despite the imperfection of the simulator, its accuracy is sufficient to train RL policies that yield useful wipes to efficiently clean the table. An extensive quantitative validation of the simulator is beyond the scope of this work and left for future research. 

\section{Conclusion}
We present a new approach to enable a mobile manipulator to wipe tables and clean spills and crumbs. The key consists of decomposing the task: we first train an RL policy to compute high-level wiping waypoints, and subsequently compute whole-body trajectories for the robot to safely execute these wipes. Our RL policy is trained entirely in simulation using an analytic SDE model of spill and crumbs dynamics and thus does not require collecting data on the robotic platform. We show that our approach outperforms wiping baselines and demonstrate our framework in hardware experiments.

\textbf{Future work}: This work opens exciting avenues of future research. First, collected data from the system could be used to train a perception module to better detect and distinguish between different types of dirty particles \cite{Yin2020}. 
Second, one could plan more complex wiping motions by combining different %
wiping primitives. 
Third, the parameters of the SDE in \eqref{eq:sde} could be learned, e.g. using neural SDEs \cite{Kidger2021,Benson2019}. They could potentially be inferred online by observing the outcome of wipes, e.g. using meta-learning approaches \cite{LewEtAl2022_rl}. 
Finally, the SDE in \eqref{eq:sde} and state representation could allow inferring and reasoning about the time-varying wiper spill-absorption properties: Since the variable $s^z$ models cleaned particles, one could keep track of the amount of wiped particles and incorporate this information for planning, perhaps by modeling the absorption parameter $\lambda$ as a stochastic process $\lambda_t$ that depends on the amount of wiped particles.

\renewcommand{\baselinestretch}{0.8932}
\bibliographystyle{IEEEtran}

\bibliography{ASL_papers,main}

\end{document}

%% file: preamble.tex
\usepackage{url}%
\usepackage{wrapfig}

\usepackage{hyperref}
\usepackage{graphicx}
\usepackage{amsmath}
\usepackage{bm}
\usepackage{bbm}
\usepackage{mathrsfs}
\usepackage{mathtools}

\let\labelindent\relax
\usepackage{enumitem}
\usepackage{cite}

\usepackage{chngcntr}
\usepackage{apptools}
\AtAppendix{\counterwithin{prop}{section}}
\AtAppendix{\counterwithin{lem}{section}}
\AtAppendix{\counterwithin{thm}{section}}

\usepackage[font={small,it}]{caption}
\usepackage{color}

\usepackage{cleveref}%

\usepackage{xspace}

\usepackage{cite}%

\usepackage{amssymb}%
\usepackage{algorithm}
\makeatletter
\renewcommand*{\ALG@name}{Alg.}
\makeatother
\usepackage[noend]{algpseudocode}

\newcommand\mydots{\hbox to 1em{.\hss.\hss.}}

\newcommand{\dd}{\textrm{d}} 
 
\newcommand{\dt}{\textrm{d}t}

\newcommand{\h}{\bm{h}}%

\newcommand{\N}{\mathbb{N}}
\newcommand{\Obs}{\mathcal{O}}

\newcommand{\A}{\mathcal{A}}

\newcommand{\U}{\mathcal{U}}

\newcommand{\E}{\mathbb{E}}
\newcommand{\R}{\mathbb{R}}

\usepackage{multirow}

\newcommand\AverageSmallMatrix[1]{{%
  \footnotesize\arraycolsep=0.75\arraycolsep\ensuremath{\begin{bmatrix}#1\end{bmatrix}}}}